\documentclass[twocolumn]{article} 
\usepackage{times}  
\usepackage{helvet}  
\usepackage{courier}  
\usepackage[hyphens]{url}  
\usepackage{graphicx} 
\usepackage[numbers]{natbib}  
\usepackage{caption} 
\usepackage{algorithm}
\usepackage{algorithmic}

\usepackage{amssymb}
\usepackage{multirow}

\usepackage{titling}
\predate{}
\postdate{}

\title{Explaining Reinforcement Learning Agents Through Counterfactual Action Outcomes}

\author {
    Yotam Amitai$^*$,
    Yael Septon$^*$,
    Ofra Amir \\
    Faculty of Data and Decision Sciences, Technion - Israel Institute of Technology\\
    yotama@campus.technion.ac.il,
    yaelfr1994@gmail.com, 
    oamir@technion.ac.il
}

\date{} 

\begin{document}

\maketitle
\def\thefootnote{*}\footnotetext{These authors contributed equally to this work.}

\begin{abstract}
Explainable reinforcement learning (XRL) methods aim to help elucidate agent policies and decision-making processes. The majority of XRL approaches focus on local explanations, seeking to shed light on the reasons an agent acts the way it does at a specific world state. While such explanations are both useful and necessary, they typically do not portray the outcomes of the agent's selected choice of action. 
In this work, we propose ``COViz'', a new local explanation method that visually compares the outcome of an agent's chosen action to a counterfactual one. In contrast to most local explanations that provide state-limited observations of the agent's motivation, our method depicts alternative trajectories the agent could have taken from the given state and their outcomes. 
We evaluated the usefulness of COViz in supporting people's understanding of agents' preferences and compare it with reward decomposition, a local explanation method that describes an agent's expected utility for different actions by decomposing it into meaningful reward types. Furthermore, we examine the complementary benefits of integrating both methods. Our results show that such integration significantly improved participants' performance.
\end{abstract}

\section{Introduction}
The behavior of reinforcement learning (RL) agents is complex - these agents operate in large, stochastic settings, and are trained through rewards received during their interaction with the environment, which is often sparse. Moreover, state-of-the-art RL models often utilize deep learning, resulting in complex models that can be hard to interpret, making the task of explaining such an agent's behavior even more complicated. While supporting people's understanding of the decision-making of RL agents is challenging, it is important for facilitating effective collaboration and appropriate trust, in particular in high-stakes domains~\cite{2210.11584} such as healthcare or transportation. 

Numerous approaches for explainable reinforcement learning (XRL) have been proposed in recent years \cite{milani2022survey}. Some methods aim to explain \emph{local} decisions of the agents in particular world-states. These include saliency maps that show what features of the state the agent attends to \cite{greydanus2017visualizing}, causal action graphs \cite{madumal2020explainable} that are based on a causal model of the agent's policy, or reward decomposition \cite{juozapaitis2019explainable} which depict the agent's expected reward for different components of the reward function, among others. In contrast to these local explanation methods, a complementary set of explanation approaches aim to convey the \emph{global} policy of the agent, that is, describe its overall strategy, behavior, or how it behaves in different regions of the state space. Such approaches include policy summaries \cite{amir2019summarizing} that demonstrate agent behavior in selected world-states based on some criteria or learning a decision tree that approximates the agent's policy \cite{liu2018toward}. 

In this work, we focus on helping laypeople understand and anticipate agent preferences by depicting the trade-offs between alternative actions. We propose a new method for visualizing the outcomes of actions \textit{not} taken by the agent as a means to explain their chosen action. We introduce ``COViz'' -- \textbf{C}ounterfactual \textbf{O}utcome \textbf{V}isualization\footnote{Code repository: https://github.com/yotamitai/COViz}, a novel local explanation method that looks beyond the agent's decision, to its manifestation.
By displaying the outcomes of both actions side-by-side, COViz provides insight into the consequences of the agent's decision. Observing the outcomes (“what if?”) can be beneficial for understanding  agent behavior or preferences. The contrastive nature of the explanation is in line with the literature on explanation in the social sciences which show that people typically provide and prefer explanations that contrast alternatives \cite{miller2019explanation}.

COViz draws inspiration from global policy summaries~\cite{amir2019summarizing}. Similar to policy summaries, COViz conveys agent behavior by showing trajectories of the agent acting in the environment. It is therefore natural to incorporate COViz in policy summaries. We propose a simple extension to COViz for generating global explanations augmented by COViz which selects summary states based on alternative action outcomes.

The idea of presenting action outcomes naturally complements many other local explanation methods that focus on state-limited reasoning. In particular, this approach pairs well with methods that present the agent’s \textit{motivation} for choosing a specific action (“why?”), such as \textit{reward decomposition}~\cite{juozapaitis2019explainable} which quantifies the agent's expectations regarding the utility of different actions. 

To evaluate the contribution of COViz to people's understanding of agent preferences, we conducted two user studies in which participants were asked to characterize the reward function of the agent by ranking its preferences. Participants were presented with explanations in the form of counterfactual outcomes (COViz), reward decomposition, or a combination of both approaches. We examined this in the context of multiple different state selection methods, based on global explanations approaches. Both studies showed that combining reward decomposition with COViz improved participants' understanding, compared to showing only one of the explanations. This result suggests that while local explanation methods are typically examined in isolation, their contributions may be complementary, and thus integrating multiple approaches has the potential to benefit users.

Our contributions are fourfold: (1) We introduce COViz, a new local explanation method for highlighting trade-offs between alternative courses of action that an agent considers; (2) We propose a way to generate global explanations augmented by COViz to incorporate alternative outcomes; (3) We integrate COViz with reward decomposition, and (4) We conduct user studies showing both that our new method is beneficial as of itself, and that its integration with reward decomposition improves users' understanding of agent behavior compared to each method separately.






\section{Related Work}
Our work draws inspiration from \emph{global} RL explanation approaches. These aim to provide users with a higher-level description of the agent by attempting to portray its behavior, strategy, capabilities, or logic ~\cite{huang2019enabling,amir2019summarizing,booth2019evaluating}. 
One approach for conveying the global behavior of an agent is \emph{Agent Strategy Summarization} ~\cite{amir2019summarizing}. In this paradigm, the agent's policy is demonstrated through a carefully selected set of world states. The strategy summarization objective is to select the subset of state-action pairs that best describes the policy of the agent. 
The criteria for selecting states can vary based on the summary objective, e.g., state importance \cite{amir2018highlights,sequeira2020interestingness,huber2021local} or using machine teaching approaches~\cite{lage2019exploring,huang2019enabling}.

Building upon this approach, the DISAGREEMENTS algorithm ~\cite{amitai2021don} portrays the diverging trajectories of two agents upon reaching a disagreement between them on how best to proceed from a given state. It provides a  side-by-side comparison of the difference in outcomes between the agents, constituting a method for agent comparison and behavior difference evaluation.
The COViz method proposed in this paper builds on and extends the DISAGREEMENTS algorithm both for the single-agent use case and for generating a \emph{local} explanation.

Local explanations in XRL seek to describe why a particular choice of action was made by a given agent in a specific world-state~\cite{tabrez2019explanation,dodson2011natural}, for example by using saliency maps to identify what elements of the environment the agent pays attention to~\cite{hilton2020understanding,huber2019,puri2020} or generating causal explanations by constructing the agent's action graph~\cite{madumal2020distal}.
~\citeauthor{juozapaitis2019explainable} (\citeyear{juozapaitis2019explainable}) use reward decomposition to provide insights into the agents' expected utility by showing users a decomposition of the environment's reward into a sum of meaningful reward types. While this approach shows what the agent expects in terms of utility, it does not show the actual outcomes of its actions. In contrast, COViz demonstrates the tradeoff between actions in terms of trajectories rather than expected reward types.

While growing, there is still an evident lack of research on how to combine different explanation methods. 
~\citeauthor{anderson2019explaining} (\citeyear{anderson2019explaining}) presents a user study that investigates the impact of explanations on non-experts' understanding of reinforcement learning agents. They investigate both a common RL visualization, saliency maps, and reward decomposition bars. They designed a four-treatment experiment to compare participants' mental models of an RL agent in a simple Real-Time Strategy game. Their results showed that the combination of both saliency and reward bars was needed to achieve a statistically significant improvement in the mental model score over the control. Other works integrated global explanations in the form of policy summaries with local explanations in the form of saliency maps~\cite{huber2021local} or reward decomposition~\cite{septon2023integrating}. These studies showed that adding reward decomposition explanations improved user performance compared to showing only summaries chosen by HIGHLIGHTS~\cite{amir2018highlights} or by frequency-based criteria.

In this work, we introduce a new local explanation method and focus on reward decomposition as a baseline for comparison, since it also illustrates tradeoffs between actions, and because saliency maps were shown to be less helpful~\cite{huber2021local}. Other approaches such as causal explanations typically require additional information (e.g., a causal graph) and cannot be implemented directly. Additionally, we show these methods can be combined and why their explanations complement each other.

\section{Background}
We assume a Markov Decision Process (MDP) setting. Formally, an MDP is defined by a tuple $<S, A, R, Tr>$ where $S$ is the set of states, $A$ is the set of actions, $R(s,a,s'): S \times A \times S \rightarrow \mathbb{R}$ is the reward received after transitioning from state $s$ to state $s'$, due to action $a$ and $Tr$ is a transition probability function $Tr(s,a,s'): S \times A \times S \rightarrow [0,1]$ defining the probability of transitioning to state $s'$ after taking action $a$ in $s$.

The Q-function is defined as the expected value of taking action $a$ in state $s$ under a policy $\pi$ throughout an infinite horizon while using a discount factor $\gamma$:
\\ $Q^{\pi}(s,a)= \mathrm{E}^{\pi}[\sum_{t=0}^{\inf}\gamma^{t}R_{t+1}| s_t=s,a_t=a]$.

\subsection{The Highway Environment}
\label{sec:Highway_env}

\begin{figure}
\centering
\includegraphics[width=0.6\linewidth]{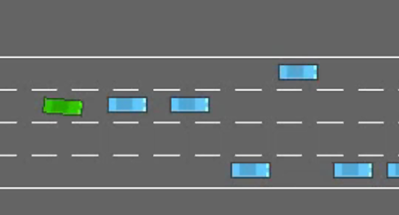}
\caption{The highway environment. The agent controls the green vehicle.}
\label{fig:highway environment}
\end{figure}

In the highway environment (Figure \ref{fig:highway environment}) the RL agent controls a green vehicle that navigates a multi-lane highway while driving alongside and interacting with other (blue) vehicles. The agent can choose to accelerate, decelerate, turn left, turn right or stay idle. The environment can be modified for the number of lanes, vehicle speed range, the number of vehicles, vehicle density, and more. Throughout the paper, we will refer to the lanes from the perspective of the agent, e.g. the bottom-most lane will be denoted as the right-most lane.

A key property of this environment is its lack of an inherent goal that must be reached, instead, an agent's reward is determined by its driving regime. This allows for designing agents with divergent, yet, reasonable strategies as no ground-truth ``optimal'' behavior exists. Additionally, the environment is both easily understandable to laypeople and represents a relatable real-world scenario.

We focus on a single domain in the current study, as both reward decomposition and approaches similar to COViz (e.g., HIGHLIGHTS, DISAGREEMENTS) have been shown to generalize to additional domains in prior work \cite{amir2018highlights,juozapaitis2019explainable,anderson2019explaining,sequeira2020interestingness,amitai2021don,septon2023integrating}.

\subsection{Reward Decomposition}
\label{sec:RD}
Hierarchical Reward Architecture (HRA) \cite{van2017hybrid} decomposes the agent's reward function into distinct components, each associated with a specific type of reward (e.g., \textit{crashing} or \textit{changing lanes} in the highway environment). This allows the Q-function to be approximated more easily by a low-dimensional representation, which can lead to faster learning. Technically, in the context of deep reinforcement learning, this architecture is implemented in a network that has multiple heads (one for each reward component) that share the lower network layers. 

\citeauthor{juozapaitis2019explainable} (\citeyear{juozapaitis2019explainable}) used this approach as the basis for a local explanation approach, called \emph{Reward Decomposition}. They present users with Reward Decomposition bars which show the Q-values for each reward component for alternative actions. This aims to reveal the agents' preferences and predictions to users. 

\section{Counterfactual Outcomes Visualization}
\label{sec:contrastive}
According to cognitive science literature, one of the key features of ``good'' explanations is that they are contrastive~\cite{miller2019explanation}.
Several XRL approaches chose to leverage some form of contrastive information in their explanation techniques \cite{silva2021teaching,lin2020contrastive,van2018contrastive}. Counterfactual explanations are a form of contrastive explanations. A counterfactual explanation provides an answer to the question ``Why $p$ rather than $q$?'', where $p$ is the fact which occurred and $q$ is some hypothetical foil which the user might have expected to occur, but did not~\cite{lipton1990contrastive}.

Many explanation techniques are not contrastive and concentrate primarily on the agent's chosen actions (\emph{fact $p$}), without considering alternative foils.
One policy-summarization (global explanation) approach that does provide contrastive information is the DISAGREEMENTS algorithm \cite{amitai2021don}. It does so by comparing two different agents to one another and visually portraying their points of disagreement.

The DISAGREEMENTS algorithm included an approach for quantifying the importance of a disagreement between the agents at a given state, by comparing their trajectories that branch out of it. The method that proved most efficient for quantifying the differences in these trajectories was comparing the values of the last state (in a limited trajectory) reached by each of them.

We build upon the contrastive nature of DISAGREEMENTS and the idea of focusing on the outcomes of alternative trajectories to devise a local explanation method. Specifically, We adapt the DISAGREEMENTS algorithm to a single agent use case and visually depict trajectories at a given state, each associated with a distinct action available to the agent. In practice, these alternative trajectories are obtained by forcing the agent to take a particular action and then allowing it to progress based on its policy for another $k$ steps. These trajectories visualize different paths the agent \emph{could have} taken (\emph{foil $q$}), had it \emph{not} chosen the specific action that it had (\emph{fact $p$}). 

We assume that the most relevant counterfactual action for the agent is its second-best choice, as this represents the most likely alternative and often provides the most informative contrast for nonexpert users. While the current focus is on this second-best action, the algorithm can be adapted through the $CF_{Meth}$ parameter (See table \ref{tb:parameters}) to consider other counterfactual options, such as the 'worst' action or a user-selected action, which we acknowledge as a potential area for future investigation.

\begin{table}[]
\resizebox{\columnwidth}{!}{%
\begin{tabular}{|c|c|l|c|}
\hline
\multirow{3}{*}{\textbf{Local}} &$k$ & \begin{tabular}[c]{@{}l@{}}Trajectory length, i.e. the number of states succeeding \\ the state where the counterfactual action was initiated.\end{tabular} & 7 \\ \cline{2-4} 
 & $Nsim$ & \begin{tabular}[c]{@{}l@{}}The number of simulations (execution traces) to run\\  for collecting trajectories.\end{tabular} & 200 \\ \cline{2-4} 
 & $CF_{Meth}$ & \begin{tabular}[c]{@{}l@{}}Counterfactual action, i.e. the method by which to\\  choose the counterfactual action.\end{tabular} & SecondBest \\ \hline
\multirow{3}{*}{\textbf{Global}} & $n$ & \begin{tabular}[c]{@{}l@{}}Summary budget, i.e. number of trajectories to\\  include in output summary.\end{tabular} & 4 \\ \cline{2-4} 
 & $overlap$ & \begin{tabular}[c]{@{}l@{}}Maximal number of shared states allowed between \\ two trajectories in the summary.\end{tabular} & 5 \\ \cline{2-4} 
 & $I_{Meth}$ & \begin{tabular}[c]{@{}l@{}}Importance method used for evaluating and ranking\\  the summary states.\end{tabular} & \begin{tabular}[c]{@{}c@{}}Last-State\end{tabular} \\ \hline
\end{tabular}%
}
\caption{COViz algorithm parameters and study values.}
\label{tb:parameters}
\end{table}

\paragraph{The COViz algorithm.} 
The pseudo-code for the algorithm is given in Algorithm \ref{alg: COViz} and parameters are summarized in Table \ref{tb:parameters} along with their user study values.

The algorithm works as follows:
Two empty lists are initialized to account for the execution traces and counterfactual trajectories (lines 3--4). For as many traces as defined, the simulation is initialized and the agent executes its policy (lines 5--22). At each simulation state $s_i$, we obtain both the agent's preferred and counterfactual action (lines 9--10). a counterfactual trajectory is obtained by having the agent initiate the counterfactual action and progress according to its policy for another $k$ steps (lines 11--16). The counterfactual trajectory is stored and the agent is reverted back to $s_i$ to progress with its preferred action (lines 17--20). Each state is added to the trace. 
Once all simulations are completed, traces and counterfactual trajectories are passed to $CFTrajPairs$  where each state is associated with both its true future trajectory and paired to the corresponding counterfactual trajectory, (line 24). The algorithm output is a set of pairs of true and counterfactual trajectories for each state in each trace.

\emph{Trajectory length $k$.}  The number of steps to include in the counterfactual trajectory ($k$) was based on the specific environment (highway) implementation, -- too few steps result in not enough time to showcase the counterfactual difference while raising the number of steps increases the influence of states succeeding the one being explained.

\emph{Complexity.} For each state reached by the agent in trace $t$ we generate a counterfactual trajectory of length $k$. Therefore the complexity is $|t| \times k$, for each trace. Importantly, these computations are done offline (once) and the choice of how many traces to simulate is configurable. 

\begin{algorithm}
    \caption{Counterfactual Outcome Visualizations}
    \label{alg: COViz}
\begin{algorithmic}[1]
    \STATE {\bfseries Input:} $\pi, k, Nsims, CF_{Meth}, I_{Meth}, n , overlap$
    \STATE {\bfseries Output:} $CO$ 
    \STATE $Traces \leftarrow$ empty list \textit{\;\;\;\#Execution traces}
    \STATE $C_T \leftarrow$ empty list \textit{\;\;\;\#Counterfactual trajectories}  
    \FOR {$i=1$ {\bfseries to} $Nsims$}
    \STATE $t \leftarrow$ empty list \textit{\;\;\;\#Current trace}
    \STATE $sim, s = InitializeSimulation()$
    \WHILE {$(!sim.ended())$} 
    \STATE $a^{\pi} \leftarrow sim.getBestAction(\pi(s))$ 
    \STATE $a^C \leftarrow sim.getCFAction(\pi(s),CF_{Meth})$ 
    \STATE $c_t \leftarrow$ empty list \textit{\;\;\;\#Counterfactual trajectory} 
    \FOR{$j=1$ {\bfseries to} $k$}
    \STATE $s^C \leftarrow sim.advanceState(a^C)$ 
    \STATE $a^C \leftarrow sim.getAction(\pi(s^C))$ 
    \STATE $c_t.add(s^C)$ 
    \ENDFOR 
    \STATE $C_T.add(c_t)$ 
    \STATE $sim, s = reloadSimulation(s)$ 
    \STATE $s \leftarrow sim.advanceState(a^{\pi})$ 
    \STATE $t.add(s)$ 
    \ENDWHILE
    \STATE $Traces.add(t)$ 
    \ENDFOR 
    \STATE $CO \leftarrow CFTrajPairs(Traces, C_T)$ 
    \STATE $\;$ \textbf{Global Extension:}
    \STATE $\mathrm{S} \leftarrow topImpTraj(CO, I_{Meth}, n , overlap)$
    \end{algorithmic}
\end{algorithm}

As opposed to DISAGREEMENTS, which only compared conflicting states between the agents, the COViz algorithm generates a counterfactual trajectory at \emph{each} step during execution. While this method does not explicitly answer the original ``why'' question, it does enable the user to implicitly infer information about the agent's preferences by its choice of action and to observe the short-term alternative outcomes of these. 

\emph{Depicting counterfactuals.} In terms of visualization, we adopt DISAGREEMENTS's visual comparison approach of displaying the counterfactual agent as a red rectangle originating around the true agent and moving away as their trajectories diverge (see Figure~\ref{fig:exp_types}). This visualization has been shown to be efficient and satisfying for users~\cite{amitai2021don}.

\paragraph{Global Extension to COViz.}
The set of counterfactual trajectory pairs obtained by COViz is stored such that a user could ask to view them for any agent-visited state. Alternatively, these can be integrated into a policy summary, showing the most meaningful pairs of counterfactual trajectories. This extension is depicted in Algorithm~\ref{alg: COViz} using the $topImpTraj$ function, (line 26).

After collecting counterfactual trajectories for each of the execution traces' states, their importance is evaluated to generate a ranking. The top-$n$ ranked states will then be used to construct the output summary. 
To determine the importance of a state $s_i$, we compare the two trajectories that branch out of it, \textit{1)} the one chosen (fact $p$) and \textit{2)} the counterfactual (foil $q$).  Importance is then calculated via the importance method $I_{Meth}$. Many criteria can be used to determine trajectory pair importance. We focus on the \emph{Last-State Importance} metric proposed in \cite{amitai2021don}, which evaluates the significance of the originating state $s_i$ based on the last state reached by the compared trajectories. 
Formally:
\[Im(s_i) = |V(s^{p}_{i+k}) - V(s^{q}_{i+k})|\]
Where $s^{p}_{i+k}, s^{q}_{i+k}$ denote the states reached by the agent following $k$ steps after selecting the fact($p$) or foil($q$) in state $s_i$ respectively.
This measure utilizes the agent's inherent value function $V(s)$ to describe the estimated utility loss of choosing the foil over the fact (i.e. optimal action). This reflects how ``far off'' from the original plan the counterfactual action has led the agent. 


The global parameter values (Table~\ref{tb:parameters}) were chosen based on pilot user studies used to design and optimize the final study. For instance, parameter $n$ which, when raised, increased study duration and cognitive load.


\paragraph{Integrating Counterfactual Outcomes with Reward Decomposition.}
COViz and reward decomposition naturally complement each other. COViz answers the question ``What if?'' by depicting alternatives, while reward decomposition answers the question ``why?'' an action was chosen based on the quantification of the expected utility. For a given state, we can integrate the approaches by showing both the reward decomposition bars for the chosen action and the counterfactual action, as well as presenting the outcome of these choices by showing the trajectories stemming from them. An example of the integration of COViz and reward decomposition can be seen in Figure~\ref{fig:exp_types}.

\section{Empirical Methodology}
To evaluate the benefits of COViz and its integration with reward decomposition, we conducted two user studies asking participants to predict preference differences between agents.
Additional experiments were conducted to compare different approaches for state selection for the studies, as further described below. 


\subsection{Agent Training} 
The agents trained for the experiment were optimized to have preference differences in their driving regimes.
Their reward functions specified positive rewards with the following behaviors: changing lanes (CL), high-speed driving (HS), and driving in the right-most lane (RML).
All agents received a negative reward for reaching a collision state. To account for each positive reward type, the number of components in the reward decomposition was set to $|C|=3$.

Three agents were trained with different preferences over the reward types (see Table 2).
No future rewards are obtained upon ending the simulation and all agents received the same negative reward for collision. We experimented with different reward-type values to get qualitatively different agent behaviors. 

Agents were trained using a double deep Q-Network (DDQN) \cite{mnih2015human} architecture for 2,000 simulations, each with a maximum of 80 steps. The network input is the observation state represented by an array of size 25 (5X5). The input layer is followed by two fully connected hidden layers of length 256. The final layer connects to three different heads, each of which is designed to account for a specific desired behavior we specified above. Each head consists of a linear layer and outputs a Q-value vector of length of 5 that predicts the Q-value for each of the possible actions: moving left (upwards), idle, moving right (downwards), going faster, going slower. 
Our implementation is based on two open-source repositories: the Highway environment and its compatible agent architectures \cite{highway-env,rl-agents}.

\subsection{State Selection}
\label{sec:state-selection}
To extract the counterfactual outcomes, we ran 200 simulations of each trained agent and saved their traces. States extracted from these traces were used in the study. The main study reported here used the last-state value method described in the COViz global extension section. 

In addition, we experimented with two alternative methods for determining state importance (parameter $I_{Meth}$). 

One approach for quantifying a state's importance is to consider the differences in Q-values between different actions~\cite{amir2018highlights}. This is usually restricted to comparing the agent's preferred action with its second-best or least preferred action. 
However, this approach proved unsuitable for our studies with COViz, as states selected using this criterion highly incentivized the agents to return to their original path after deviating with a single action, rendering the counterfactual trajectories almost identical to the true ones. We note this effect is domain-dependent -- in some domains it is easy to return to the original policy after deviating, while in other domains it may not be possible.


Another approach for selecting states is sampling them based on the frequency of the agent encountering the states, thus reflecting the agent's ``typical'' behavior. Technically, this can be done by randomly sampling states from execution traces. We ran a replication of our main study with this approach (See supplementary material\footnote{Supplementary material can be found at \url{https://t.ly/AhrTf}} for complete information, and a short summary in the Results section). 


\begin{table}
    \centering
    \begin{tabular}{|c|c c c c|}
        \hline
        Reward Type  & CL  & HS & RML & Collision  \\
        \hline
        Agent 1 & 3 & 1 & 8 & -3 \\
        Agent 2 & 5 & 8 & 1 & -3 \\
        Agent 3 & 8 & 1 & 5 & -3\\
        \hline
        
    \end{tabular}
    \caption{User-study agents' reward-type values.}
    \label{tab:reward setting}
\end{table}

\begin{figure}
\centering
\includegraphics[width=0.7\linewidth]{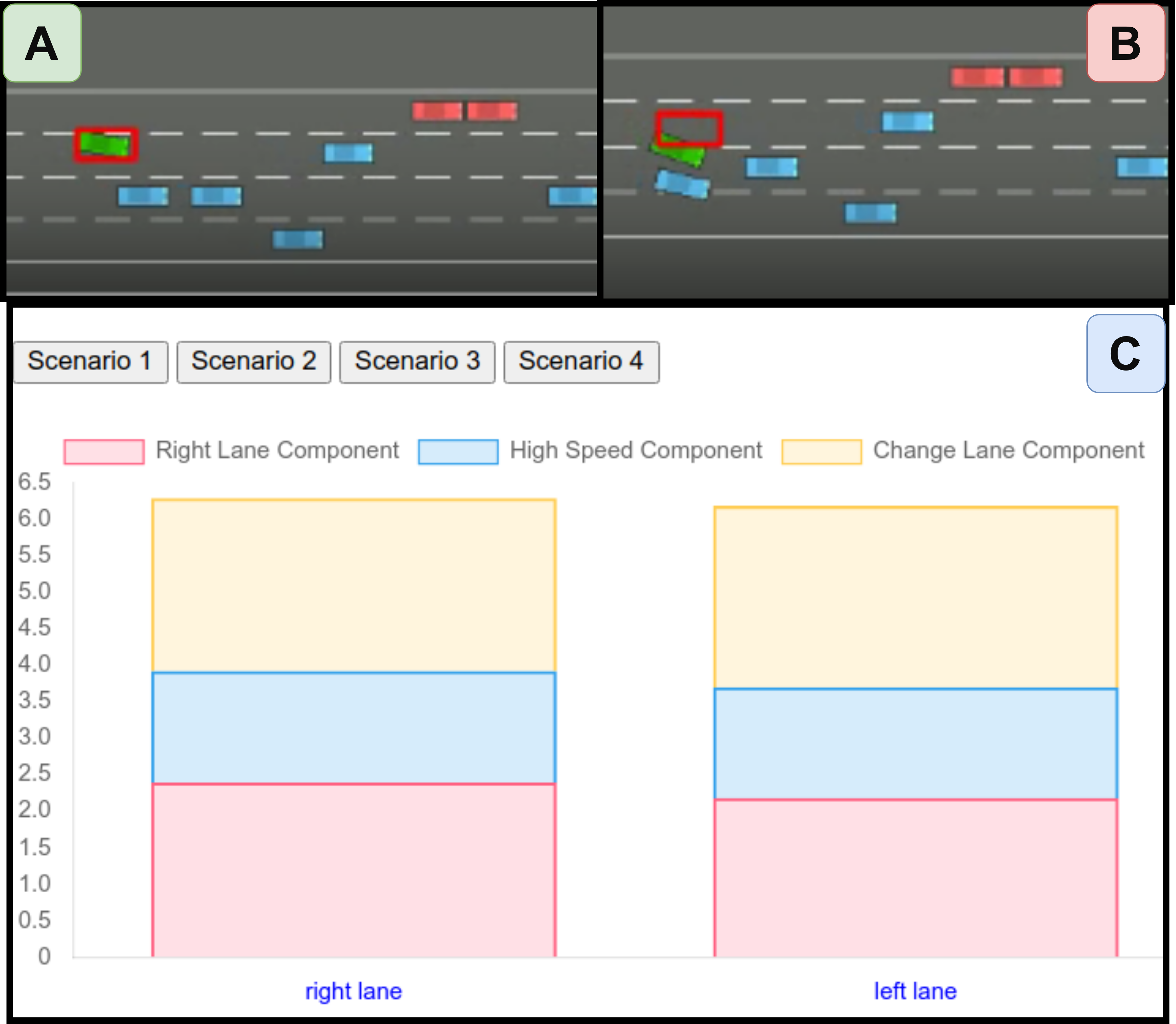}
\caption{User study explanation types. \textbf{A+B. Counterfactual Outcomes (COViz)}: videos depicting the different outcomes of counterfactual actions. A and B depict the start and end frame of the video respectively; \textbf{C. Reward Decomposition (RD)}: a graph presenting the value each reward component contributed to the chosen action at the portrayed state. \textbf{Integrated method (CORD)}: The COViz outcome video and RD graph together. RD graph portrays both the true and counterfactual action.
}
\label{fig:exp_types}
\end{figure}

\subsection{Study Design}
We conducted two studies to evaluate the contribution of COViz and its integration with reward decomposition to people's understanding of agents' preferences. Study 1 was a between-subject study, where participants were exposed to a single type of explanation. This design avoids learning effects but suffers from higher variance due to differences between study participants and does not allow to elicit users' subjective preferences. To address this limitation, study 2 used a within-subject design where participants were exposed to all explanation methods, and could thus compare them directly. Both studies were approved by our institutional review board.

\emph{Experimental conditions}. The studies included three local explanation types: Counterfactual outcomes (\textit{COViz}), reward decomposition (\textit{RD}), and their integration (\textit{CORD}). Examples of explanation types can be seen in Figure \ref{fig:exp_types}. Additionally, a control condition of summaries without local explanations (\textit{SUM}) was tested in study 1.
In study 1, participants were assigned a condition and observed all agents using their assigned explanation. 
In study 2, participants saw each agent with a different, randomly assigned, explanation type. Since it focused on participants' preferences for local explanations, we did not include the summary-only control in this study. The ordering of the \textit{COViz} and \textit{RD} explanations was random, however, the combined method was always shown last as it was more natural to show it after each method appeared separately. While this may introduce a learning effect, study 1 does not suffer from this confound.


\emph{Task.} 
In both studies, participants' task was to assess the behavioral preferences of the different agents. For each agent, participants observed four scenarios (i.e., states selected as explained above) and their explanations. 
Since the Q-values of each agent are different, the ``important'' states differ as well. However, for a specific agent, all participants viewed the same scenarios. After participants observed the explanation of the agent's behavior in each scenario, they were asked to predict the agent’s preferences between two reward components, e.g. ''based on the video, which component do you think the agent (green car) cares about more, high-speed or lane change?'' (Full surveys in supplementary material). Participants could return and review the explanations while answering the questions. The agents' order and the task answers were randomly presented. 

After each task, participants rated their agreement on a 7-point Likert scale \cite{likert1932technique} with statements adapted from Hoffman's explanation satisfaction questionnaire \cite{hoffman2018metrics}. These adaptations included specifying the task at hand rather than using general terms like 'goal' or 'task', and removing items irrelevant to our study objectives (see supplementary for full scale). Notably, despite these minor modifications, the internal consistency of our adapted scale was confirmed, yielding high Cronbach's Alpha values ($>$0.9) across all studies.

Then, participants were asked to rate their confidence in each answer on a Likert scale from 1 (``not confident at all'') to 5 (``very confident'') and they had the option to describe their reasoning in a free-text response.
In study 2, where participants saw multiple explanation types, they were also asked to rank their preferences among them and provide textual feedback regarding how each method helped them.

\emph{Procedure.} In both studies, participants were first introduced to the highway environment. Then, participants learned about their condition's explanation method along with an example to assist in interpreting the visualized information. In study 1, participants were presented with only one of the four explanation types: \textit{Sum}, \textit{COViz}, \textit{RD}, or \textit{CORD}. The explanation type was randomly assigned and consistent across all three tasks. In study 2, before each task, instructions and information regarding the upcoming explanation method were provided, after which, a demonstration for the task they would be asked to complete was shown. Prior to starting the task itself, participants were quizzed on their understanding and only allowed to proceed once answering all questions correctly. Each task included an attention check question. Finally, participants were asked to provide demographic information and rate their proficiency in AI. All study materials including the specific instructions for each explanation method can be found in the supplementary.

\emph{Participants}. We recruited participants through Amazon Mechanical Turk ($N_{1}=147$ and $N_{2}=50$ for study 1 and study 2, respectively). On average, study participants self-proclaimed low proficiency in AI which aligns with the purpose of explanations for laypeople. Participants were native English speakers, ages 18 - 55 from the US, UK, or Canada.
Participants received a \$3 base payment and an additional bonus 30-cent bonus for identifying the preferences of each of the agents correctly. Participants who failed the attention question were excluded from the final analysis, as well as participants who completed the survey in less time than two standard deviations of the mean completion time. After exclusions, we had 115 participants (SUM=27, COViz=30, RD=31, CORD=27) in study 1 and 49 participants in study 2 with a mean age of 39 \& 37, and a female count of 39 \& 27 for study 1 and 2 respectively.
We note that the number of participants in each experiment aligns with previous user studies in XRL, e.g. \cite{huang2018establishing,sreedharan2020tldr}.


\section{Results}
The objective measure used in our analysis is the mean fraction of correct reward component comparisons, i.e., participants' correctness rate, for each condition. We used the non-parametric Mann-Whitney $U$ test in the statistical analyses. We note that there were no differences between different demographic groups (age or gender).

Overall, the integration of the two explanation types improved participants' ability to asses the agents' preferences. 
These results were replicated in both studies.
In study 1, the integration method of combining both explanation types CORD (\textit{M}= 0.64) led to significantly improved performance compared to  RD or COViz alone (RD: M=0.45, RD vs. CORD: $U=204,p<0.001$; COViz: M=0.42, COViz vs. CORD: $U=657.5, p<0.001$). Comparing to SUM (\textit{M}= 0.31) there was a significant improvement in the performance as well ($U=607, p=0.002$). The difference between COViz and RD was not significant ($p = 0.22$).
Similarly, in study 2, the combination of explanation types, CORD (\textit{M}= 0.63), significantly improved participants' performance compared to RD  or COViz (RD: M=0.48,  RD vs. CORD: $U= 767, p<0.001$ and COViz: M= 0.4,  COViz vs. CORD: $U= 1814.5, p<0.001$).  Here too, the difference between COViz and RD was not significant ($p = 0.064$). 

Moreover, in our experimentation with frequency-based state selection, we found that all local explanation methods were useful, with no significant difference between them, when addressing informative states. However, for less informative states all explanation methods proved less useful than a coin toss, except the integrated approach which was still superior (see supplementary material).

While CORD participants were subjected to more information than the other conditions, prior work has demonstrated that more information does not necessarily lead to better outcomes~\cite{anderson2019explaining}, implying that integration between these methods successfully captures relevant and complementary information for performance without imposing excessive cognitive load.

\begin{figure}
\begin{minipage}{0.49\linewidth}
\centering
\includegraphics[width=\linewidth]{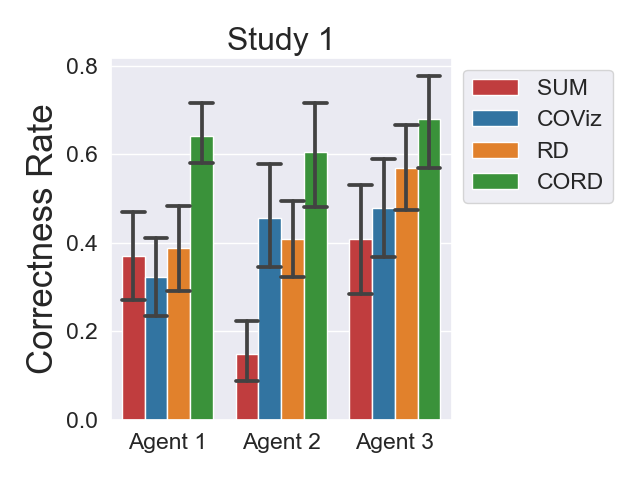}
(A)
\end{minipage}
\begin{minipage}{0.49\linewidth}
\centering
\includegraphics[width=\linewidth]{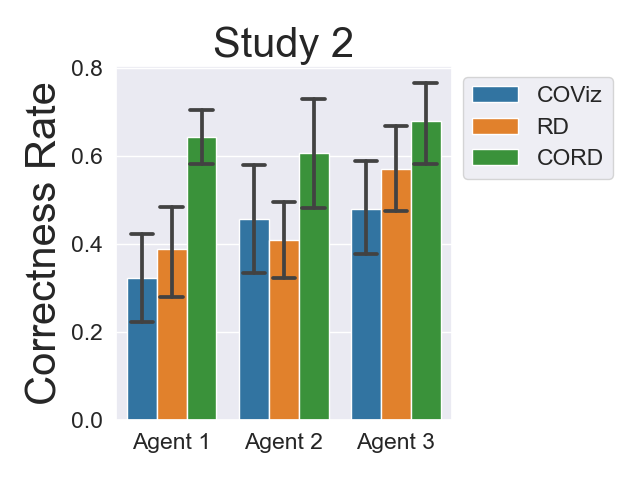}
(B)
\end{minipage}
\caption{Participants' mean success rate in identifying the preferences by condition and agent in study 1 (A) and study 2 (B). The error bars show the 95\% CI.}
\label{fig:first_study_agent1}
\end{figure}

In study 1, when breaking down the results by agent, the trend was consistent with the overall results. The greatest improvement from the integration of the two approaches was for Agent 1 (Figure~\ref{fig:first_study_agent1}). We note that Agent 1 is a tricky agent as its second preference - being in the right-most lane is correlated with its first preference changing lanes. This shows that the combination of the two explanations has a significant impact, especially for challenging behaviors. Participants' comments also mentioned the benefit of combining the different explanations. For instance, one participant wrote ``The two [explanations] together helped me understand but separately I was a lot more confused''. Another comment was ``The graphs help to visualize how the agent is rewarded for behaviors while the videos show the results of rewarding (or prioritizing) those actions for the agent.''

\begin{figure}
\centering
\includegraphics[width=0.5\linewidth]{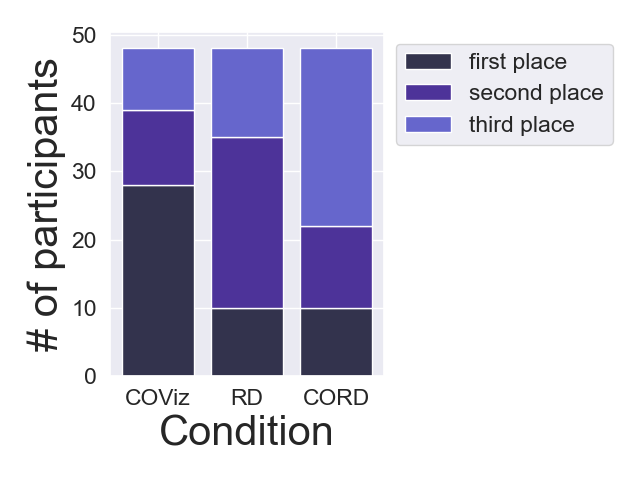}
\caption{The distribution of participants' ranking of the three explanation types}
\label{fig:ranking}
\end{figure}

In study 2, we asked participants to rank the three explanation types.
Participants differed in their preferences. Some preferred the video-based approach of COViz, e.g., ``video is always easy for me to comprehend'', while others preferred the reward bars: ``The graph is more efficient''. A participant that preferred the combination wrote: ``It is more instructive to both see the component values and then watch the video. ''
The overall ranking results show that COViz was ranked first by 28  participants ($57\%$ of participants), while the combination of the two explanation types was ranked in last place by 26 participants  ($53\%$ of participants) as shown in Figure~\ref{fig:ranking}.

For both studies, participants' confidence and satisfaction ratings were above the neutral rating ($>3$). Specifically, for study 1 and study 2 the satisfaction results for each of the conditions were COViz: M=4.9 and 5.16, RD: M=5.06 and 5.06, CORD: M=5.29 and 5.04 respectively and SUM=5.4 in study 1 (Graphs provided in supplementary). With respect to participants' confidence, the ratings by condition observed in studies 1 and 2 were COViz: M=4.04 and 3.93, RD: M=3.8 and 4.02, CORD: M=4.03 and 3.83, respectively). None of the differences between conditions in terms of satisfaction or confidence were significant.

\section{Discussion and Future Work}
This paper presented a new method for explaining RL agents through counterfactual outcomes, and the benefits of combining them with another local explanation - reward decomposition. We conducted user studies to evaluate the contribution of this approach to people's ability to analyze agent preferences as well as rating the participants' preferences of each explanation type. Our results show that both explanation types, COViz, and reward decomposition, resulted in better-than-random participant performance, but the integration of the two enabled participants to reach a significantly higher success rate. It is not trivial that this would be the case since the benefit might only stem from one explanation, and the cognitive load might be higher as was observed in some studies~\cite{anderson2020mental}. Indeed, when showing frequency-based chosen states, COViz or reward decomposition alone did not always result in higher performance than random guess. However, showing both types of explanations together did improve performance beyond random guessing.

Even though the combination of the explanation types led to a higher correctness rate, when participants were asked to rate the explanation types, the combination was ranked last. We hypothesize that the reason for that is the mental overload on the participants. For example, one participant commented that ``...the combination is too confusing for me.''
When combining two explanation types participants are required to pay attention to more details. This finding is in line with previous work stating that subjective satisfaction and proxy tasks often do not align with objective performance ~\cite{buccinca2020proxy}.
A potential future-work solution is to reduce user cognitive load by cascading the information and allowing users to control the amount they view at once, thus avoiding overwhelming them and engaging them more.


Our experimentation of state-selection criteria highlighted a vulnerability of COViz -- only a single counterfactual action is constrained. In the generalized continuous setting, a single action deviation does not necessarily result in a divergent outcome and is more likely to be negligible or easily reversible. Future work can alleviate this limitation by enforcing an alternative sequence of actions rather than forcing only a single deviation from the policy.

In this work, the selection of counterfactual actions was guided by heuristic methods. Alternatively, this choice can be handed over to users to enhance interactivity and foster trust, or for purposes of debugging.

Finally, while this work focused on the task of identifying agent preferences, future work can try to characterize which explanation types are most useful for different user tasks, and explore the combination of other local explanation types, including saliency maps, action graphs, and others.

\section{Acknowledgements}
The research was partially funded by the European Research Council (grant \texttt{\#101078158} - CONVEY), and by the Israel Science Foundation (grant no. 2185/20). The authors thank Tobias Huber for helpful advice and feedback.

\bibliographystyle{IEEEtranN}
\bibliography{camera_ready}

\end{document}